\documentclass[10pt,a4paper,conference]{IEEEtran}
\usepackage{caption}

\ifCLASSINFOpdf
  \usepackage[pdftex]{graphicx}
  \graphicspath{{img/}}
   \DeclareGraphicsExtensions{.pdf,.jpeg,.png}
\else
\fi
\usepackage{color}
\usepackage[table,xcdraw]{xcolor}
\begin{document}
%
\title{Learning-based Tracking\\ of Fast Moving Objects}

\author{

\IEEEauthorblockN{Ale\v{s} Zita}
\IEEEauthorblockA{Institute of Information Theory \\ and Automation of the\\ Czech Academy of Sciences\\
Pod Vodárenskou věží 4\\
Email: http://www.utia.cas.cz/people/zita}
\and
\IEEEauthorblockN{Filip \v{S}roubek}
\IEEEauthorblockA{Institute of Information Theory \\ and Automation of the\\ Czech Academy of Sciences\\
Pod Vodárenskou věží 4\\
Email: http://www.utia.cas.cz/people/sroubek
}}

\maketitle

\begin{abstract}
Tracking fast moving objects, which appear as blurred streaks in video sequences, is a difficult task for standard trackers as the object position does not overlap in consecutive video frames and texture information of the objects is blurred. Up-to-date approaches tuned for this task are based on background subtraction with static background and slow deblurring algorithms. In this paper, we present a tracking-by-segmentation approach implemented using state-of-the-art deep learning methods that performs near-realtime tracking on real-world video sequences. We implemented a physically plausible FMO sequence generator to be a robust foundation for our training pipeline and demonstrate the ease of fast generator and network adaptation for different FMO scenarios in terms of foreground variations.
\end{abstract}


\begin{figure*}[!ht]
  \centering
  \includegraphics[width=0.45\textwidth]{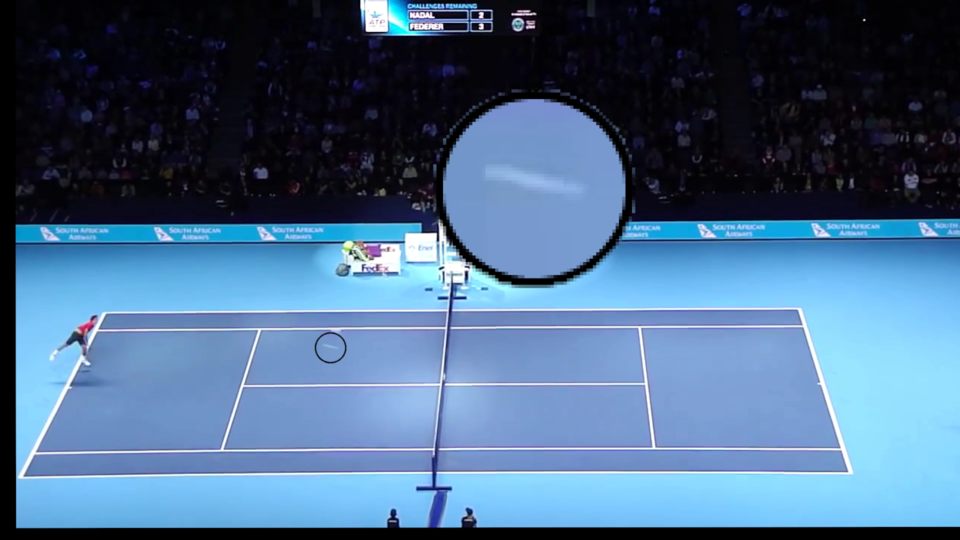}
  \includegraphics[width=0.45\textwidth]{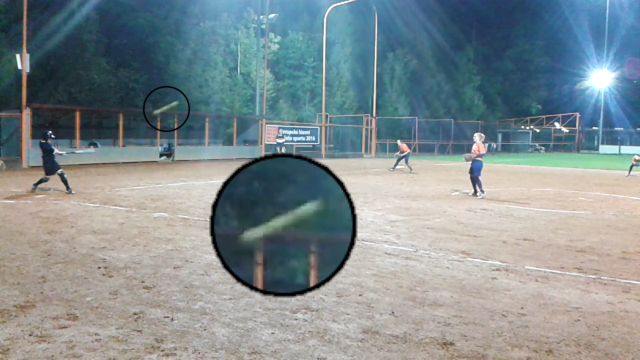}
  \includegraphics[width=0.32\textwidth]{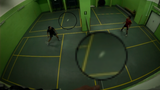}
  \includegraphics[width=0.32\textwidth]{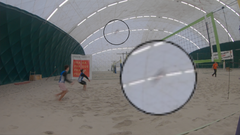}
  \includegraphics[width=0.32\textwidth]{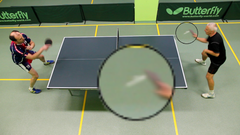}
\caption{Examples of Fast Moving Objects in real-world videos. }
\label{fig:FMO_examples}
\end{figure*}

%
\IEEEpeerreviewmaketitle


\section{Introduction}
\label{sec:intro}




Object tracking is a well explored field of computer vision. The majority of object tracking algorithms starting from basic correlation trackers up to state-of-the-art deep network trackers utilize texture-based correlation or feature based methods. Modern video capturing devices with built-in processing algorithms are capable of producing sharp images of the moving objects. Moreover, the person capturing the object in motion typically tracks the moving object, hence it predominantly stays in the center of the image and in-focus. For such tasks the correlation-based trackers are thus sufficient. 

The situation dramatically changes when the object in question moves so fast, that it is impossible to capture it sharp in videos. We call such object in motion an \emph{'FMO'}, short for \emph{Fast Moving Object}~\cite{rozumnyi2017world}. 

An FMO can be loosely defined as an object traveling a distance larger than its diameter within one frame of the video sequence (Figure~\ref{fig:FMO_examples}). The inter-frame object overlap is negligible and this causes problems to many conventional trackers. 

A typical manifestation of an FMO in video frames is a prolonged streak without any particular texture, colored with prevailing color of the object, or a combination of object colors; see Figure~\ref{fig:FMO_examples}. The lack of any sharp texture of the object renders most of the texture-based correlation trackers unusable. 

The first tacking algorithm specifically designed for FMOs uses a method based on background subtraction~\cite{rozumnyi2017world}. This technique requires static background, static camera and large prominent foregrounds. It is also prone to object miss-tracking, which then requires a time consuming object re-detection. 

More recent approaches deal with the problem of FMO tracking by running a de-blurring algorithm~\cite{rozumnyi2019non,kotera2018motion,kotera2019intra}. These methods perform considerably better, but are extremely slow, as they require a full-blown de-blurring optimization pipeline. Thereforhe they are unusable for real-time video stream processing.

The proposed solution is to approach the problem not as tracking by correlation but rather as tracking by segmentation. Our view is that the segmentation task is for this cases more useful. Resulting segmentation can be further used for trajectory prediction and down the pipeline even for the trajectory estimation in the conventional de-blurring algorithms. 

Our primary goal is to provide a method operating in real-world scenarios such as tracking of ping pong, squash balls, badminton and similar objects. The method uses a convolutional neural network (CNN) with real-time performance in videos with resolution 320x240. For training and evaluation, we synthesized a dataset composed of relevant YouTube sport video sequences. We propose on-demand synthetic FMO data generator to tackle the problem of producing annotated data automatically. 

For comparison with state-of-the-art approaches we evaluate our method on a well established FMO dataset \cite{rozumnyi2017world}. We demonstrate that our approach shows competitive results and investigate cases where our algorithm outperforms and under-performs current methods both in precision and execution time.



\section{Related Work}

\textbf{Video Object Tracking}

Object tracking is well established field of research in computer vision. Many methods have been proposed for tracking single or multiple objects in video sequences. Namely tracking by detection\cite{hare2015struck,zhang2014meem}, tracking by features\cite{chen2010multi,gladh2016deep} tracking by correlation\cite{liu2018long} and others. All of the mentioned approaches are based on either object detection using texture information of the tracked object or features extracted from it. This assumes that the object image contains some minimum level of details. Also, many of the conventional trackers performs best when the tracked object bounding boxes largely overlap in the consecutive frames. Both the mentioned assumptions does not hold in sequences with a FMO.



\textbf{FMO Tracking}

FMO tracking is bringing attention of more and more researchers lately. 
Initial work in this field was done in \cite{rozumnyi2017world}, where the authors firstly introduced the theme and proposed first tracker based on background subtraction. In the heart of the method lies a tracker capable of tracking the background changes. When the tracker fails a time consuming re-detection is ran to resume tracking.
Lately, an interesting work was done in \cite{rozumnyi2019non} where the tracking problem was defined as a de-bluring optimization problem. In another similar approach \cite{kotera2019intra} authors show intra-frame tracking capability of de-bluring approach. Albeit the results are promising in both mentioned publications, these methods focus on videos with static camera and background and additionally, their algorithm cannot be used real-time due to high processor time demands of the optimization algorithm.

\begin{figure*}[!t]
  \centering
  \includegraphics[width=0.95\textwidth]{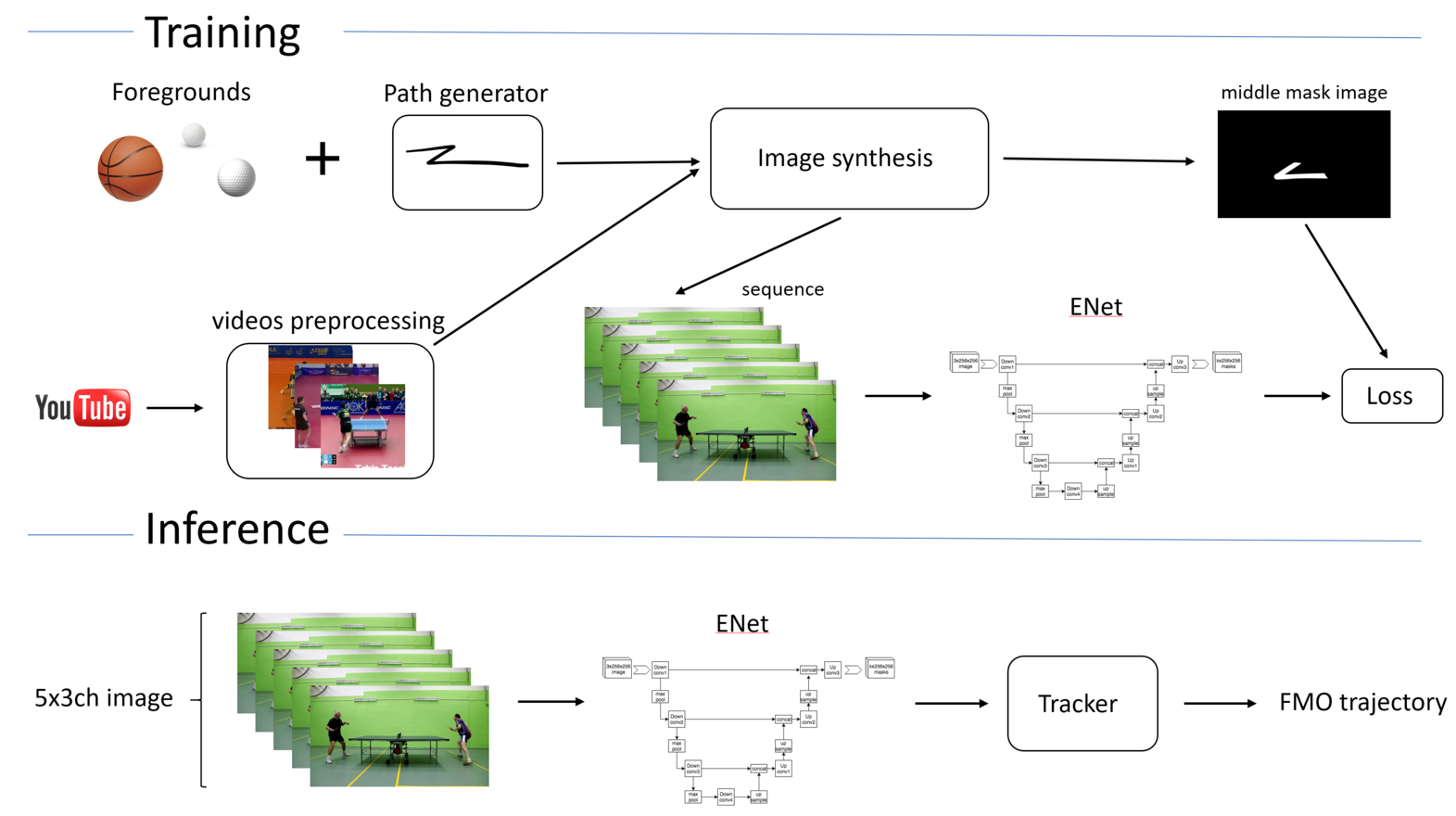}
\caption{Processing pipeline: During the training phase (top section) the sequences are dynamically synthesized using pre-processed video sequences, foregrounds and path generator. Next, the frames are concatenated and input the network as a 15-channel image. During inference phase (bottom part) the sequences are segmented by the network and Kalman based tracker is used for path prediction. }
\label{fig:NET}
\end{figure*}

\section{Method}
\label{sec:method}

In this section, we first give a brief introduction of the
overall framework of the proposed method. Then we briefly investigate various strategies and finally we described the proposed method in depth.

\subsection{Overview}
The work of \cite{rozumnyi2017world} inspired us to tackle the problematic cases on which the method did not perform well enough, namely tracking of the very small objects. Most of the sport videos are sharp by itself from using modern capturing devices or because the cameraman is actively tracking the object of interest. An exception of this are small objects moving very fast. Those are typically balls in some specific sports such as tennis, softball or badminton.

After some failed attempts to solve tracking of such small fast objects (FMO) by conventional means, we have turned our attention towards deep learning methods.
Deep learning methods achieve top results in many segmentation tasks in terms of both the computation time and the precision. We researched several state-of-the-art segmentation networks of which we were able to achieve the best results with u-net type architecture with inception bottleneck modules called ENet\cite{paszke2016enet}. Please refer to Figure \ref{fig:NET}.




Because we aimed to be able to track problematic real-world sports videos where due to the bad resolution and small object other tracking methods fail, we included YouTube sports videos for performance evaluation. However the used YouTube dataset is not annotated and was evaluated only by human ratter's personal opinion. Therefore it is not included in resulting statistics. This dataset had however profound influence on network creation and hyper-parameter tuning. 

As a benchmark dataset we choose the publicly available FMO dataset to be able to compare the performance of proposed approach with the original method\cite{rozumnyi2017world}. We perform preprocessing of the dataset in such a way that the foregrounds size and color resembles the foregrounds used for training. Another way to make the method perform on different foregrounds is to fine-tune the network with different dataset generator parameters. Example of fine-tuning can be seen in Table \ref{tab:results} method II where we have fine-tuned the network to detect bigger foregrounds, e.g. frisbee or volleyball.

\subsection{Backbone architecture}
As in many cases of most modern methods, our approach is based on deep network architecture. 
The backbone of our segmentation network we use U-Net architecture called ENet\cite{paszke2016enet}, consisting of inception blocks based on \cite{szegedy2016rethinking}. The initial choice of this 
network design was done based on the speed of evaluation and performance on various benchmark datasets. We found out that this design outperformed other backbone designs we had tested, such as classical U-Net or Mask-RCNN.

The basic idea behind the FMO trace segmentation is training the network to recognize prolonged objects with no apparent texture, typically of white color to resemble most common sports balls, such as in ping pong, squash and even badminton or tennis. This represented in our opinion majority of the problematic sport videos. 

Furthermore, because of the difficulty of the task, the single image segmentation proven to be to difficult and produced too many false positives. This is expected, as the proposed network basically learns to recognize bright smears and therefore falsely segment any bright spots or lines in the image. To overcome this, we tested approaches with sequence of several consequent frames as a network input. The idea was, that sequence of images improves trace consistency in time. For this purpose, we tested several multi-frame approaches, namely 3 and 5 frames either concatenated in color channels or as a full 4D input to the 4D network (Even though this approach is mathematically equivalent to the channel concatenation the idea was to produce faster learning and less false positives.)  
The best results were achieved by using 5 consequent video frames concatenated in color channel, i.e. the input to the network is single 15 channel image. See Figure \ref{fig:NET}

The images used for training are synthetic FMO sequences based on real-world sporting background images. Because every deep network is only as good as the dataset used for training, we have invested considerable effort to create a quality tool for generation the synthetic sequences. Please refer to the Section \ref{fig:FMO_generator} for further details.

Although majority of the state-of-the-art deep learning methods heavily depends on the re-using of the learned parameters from their successful predecessors, transfer learning proven inapplicable in our case. This is due to the specificity of our task, which cannot exploit learned convolution kernels from other problems based on extraction of texture features.

\subsection{Dataset generator}
\label{sec:datagenerator}

\begin{figure*}[t]
  \centering
  \includegraphics[width=0.32\textwidth]{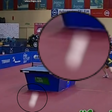}
  \includegraphics[width=0.32\textwidth]{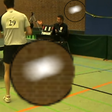}
  \includegraphics[width=0.32\textwidth]{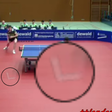}
\caption{FMO synthetic data generator example. The rightmost image shows example of small emulated bounce.}
\label{fig:FMO_generator}
\end{figure*}

In the heart of any modern machine learning is always a good dataset. Due to the nonexistence of any training FMO dataset, we created our own physic laws obeying  FMO sequence generator. First we have obtained the dataset of youtube sports videos which we used as a background. To eliminate any false fast moving object from the videos, we have generated sequences of median images. Every frame of such a sequence was calculated as a median of 5 consequent frames. Next, we created a foreground generator based upon selected ball images from variety of sports. Finally, we designed physically plausible generator of trajectories, including random bounces or occlusions. See Figure \ref{fig:NET}.

In the core of the image synthesizer is a random motion path generator which takes into account fully simulated camera (including CCD size resolution and aperture properties) as well as motion of the simulated object in space. The generator begins from a random initial speed vector and then iterates in time simulating the motion. For even better plausibility the gravitational acceleration into account, too. The sudden velocity changes (hit from a racket like), bounces (as if from wall, ground or table) occlusions and sudden motion stops are simulated as well. Refer to Figure \ref{fig:FMO_generator} right. 

Such a generated trajectory is then convolved with foreground to create the motion trace and finally inserted as a weighted sum into the sequence of background images using following formula. 

\begin{equation}
\label{eq:image_synthesis}
I_t(x) = [P_t * b_f F](x) + (1-[P_t * M])B(x),
\end{equation}

where $P_t$ is the path PSF normalized to sum to 1, $F(X)$ is the random foreground image, $b_f$ is the overexposure brightness factor (described in next paragraph), $M(x)$ is foreground indicator function and $B(x)$ is the background image.
Used foreground image is created as a random selection of real-world white ball images which are tinted in random bright color and resized to a pre-defined range of foreground sizes. 

Another aspect which had been taken into account is fast moving object overexposure. This is due to to the 'HDR' effect of the moving object. The overall brightness of the object in one frame can, and often is, brighter than maximum brightness point in the rest of the image. Typically what every camera has to solve is the conversion of high brightness range of the world to the quantized 255 brightness values. This is done by several techniques which are out of scope of this article. This conversion usually includes some form of clipping of the brightness levels which are too high to optimize overall image brightness balance. In a typical image without any FMO the overexposed parts of the image are clipped to the maximum allowed brightness. But, in case of a fast moving object, the true brightness of the object when stopped is an integration of its brightness along the object trajectory. In other words, the overall brightness of the object is spread out along the object path so it does not exceeds the maximum pixel brightness in any point of the image. Therefore, it is often the case, that the true brightness of the object, when aggregated along the path, exceeds the maximum brightness of the image, especially with the white ball. If this effect would not have be taken into the consideration, the rendered object would seem very dim in the resulting image. This lead us to set the factor of absolute brightness of foreground between 0.8 - 1.4 of the maximum brightness.

As for ground truth mask image used in training phase, we use the foreground path mask corresponding to the middle frame of the sequence. It is calculated again as foreground mask convolved with the trajectory corresponding to the middle frame ($[P_3 * M]$). Please see the Figure \ref{fig:NET} for illustration.

\subsection{Tracking}
\label{sec:tracking}
On top of the successful segmentation we have implemented a simple tracker. The tracker is responsible for final object trajectory estimation. First we select the blob which most likely represent the tracking object. This can be achieved by simply selecting the largest connected component in segmentation image. For sequences containing many false positives, more sophisticated logic can be applied. We used weighted composition of two measures: connected component size and shape. Since we are looking for prolonged object, we use second central moments of the connected components to estimate the prolongation. The position of the blob is in tracker represented by bounding box.

Sequences of the bounding box positions are used by the tracker to extrapolate the object trajectory. For frames with missing or too small blobs, we utilize a Kalman filter to estimate missing trajectory or predicting trajectories in cases the object is lost or occluded. 

\begin{figure}[t]
  \centering
  \includegraphics[width=0.45\textwidth]{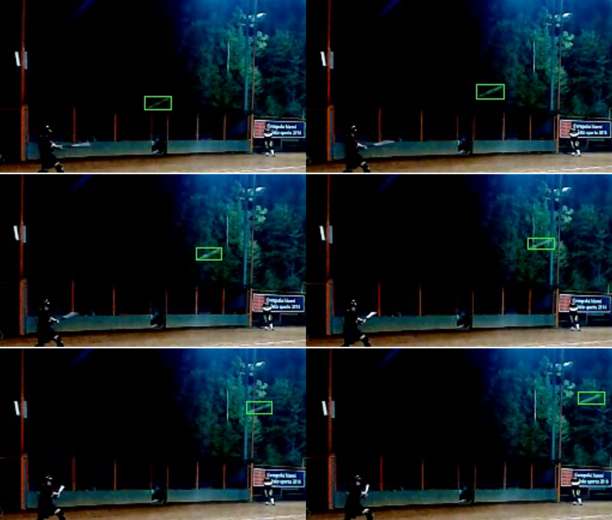}
\caption{Tracker output example}
\label{fig:tracker}
\end{figure}

Output of the tracker is a sequence of bounding boxes representing estimated object trajectory. Refer to examples in Figure \ref{fig:tracker}

\section{Experiments}
In this section we present results of proposed method and compare them to the original CVPR paper\cite{rozumnyi2017world}. We focused our attention at real-world application with both the speed of the inference and the accuracy for small ball-like object detection.

\subsection{Evaluation}

The proposed method was evaluated on the FMO dataset \cite{rozumnyi2017world}, where it achieved comparable or better results when compared to the published method. 

The performance criteria was selected to correspond evaluation statistics in the original paper. These are precision TP/(TP + FP), recall TP/(TP + FN) and F1-score 2TP/(2TP + FN + FP), where TP, FP, FN is the number of true positives, false positive and false negatives, respectively. A true positive detection has an intersection over union (IoU) with the ground truth polygon greater than 0.5 and an IoU larger than other detections. The second condition ensures that multiple detections of the same object generates only one TP. False negatives are FMOs in the ground truth with no associated FP detection.

The results for both the original method and our approach are listed in Table \ref{tab:results}. It can be concluded, that overall mean F1-score is slightly better for our method, as well as mean recall. We were also able to avoid significant under-sizing of the resulted segmentation of the FMO trace, which would cause high precision values over small recall value. Therefore, we argue that our approach results are more balanced in terms of precision and recall performance metrics.

Training of the segmentation network was performed using the synthetic data generator described in Section \ref{sec:datagenerator}.

\begin{table*}[]
\captionsetup{font=scriptsize}
\centering{
\begin{tabular}{
>{\columncolor[HTML]{C0C0C0}}l llllllllllll}
\cellcolor[HTML]{FFFFFF}                                         &                                                  & \multicolumn{3}{c}{\textbf{original work}}                                                                                                             &                          & \multicolumn{3}{c}{\textbf{Learning-based I}}                                                                                                                                            &                       & \multicolumn{3}{l}{\textbf{Learning-based II}}                                                                                                             \\ \cline{1-5} \cline{7-9} \cline{11-13} 
\multicolumn{1}{|l|}{\cellcolor[HTML]{C0C0C0}}                   & \multicolumn{1}{l|}{\cellcolor[HTML]{C0C0C0}n}   & \multicolumn{1}{l|}{\cellcolor[HTML]{C0C0C0}Pr.} & \multicolumn{1}{l|}{\cellcolor[HTML]{C0C0C0}Rec.} & \multicolumn{1}{l|}{\cellcolor[HTML]{C0C0C0}F1} & \multicolumn{1}{l|}{}    & \multicolumn{1}{l|}{\cellcolor[HTML]{C0C0C0}Pr.}           & \multicolumn{1}{l|}{\cellcolor[HTML]{C0C0C0}Rec.}          & \multicolumn{1}{l|}{\cellcolor[HTML]{C0C0C0}F1}            & \multicolumn{1}{l|}{} & \multicolumn{1}{l|}{\cellcolor[HTML]{C0C0C0}Pr.} & \multicolumn{1}{l|}{\cellcolor[HTML]{C0C0C0}Rec.} & \multicolumn{1}{l|}{\cellcolor[HTML]{C0C0C0}F1} \\ \cline{1-5} \cline{7-9} \cline{11-13} 
\multicolumn{1}{|l|}{\cellcolor[HTML]{C0C0C0}volleyball}         & \multicolumn{1}{l|}{\cellcolor[HTML]{EFEFEF}50}  & \multicolumn{1}{l|}{100}                         & \multicolumn{1}{l|}{45.5}                         & \multicolumn{1}{l|}{62.5}                       & \multicolumn{1}{l|}{}    & \multicolumn{1}{l|}{\cellcolor[HTML]{EFEFEF}0}             & \multicolumn{1}{l|}{\cellcolor[HTML]{EFEFEF}0}             & \multicolumn{1}{l|}{\cellcolor[HTML]{EFEFEF}0}             & \multicolumn{1}{l|}{} & \multicolumn{1}{l|}{33.3}                        & \multicolumn{1}{l|}{42.9}                         & \multicolumn{1}{l|}{37.5}                       \\ \cline{1-5} \cline{7-9} \cline{11-13} 
\multicolumn{1}{|l|}{\cellcolor[HTML]{C0C0C0}volleyball passing} & \multicolumn{1}{l|}{\cellcolor[HTML]{EFEFEF}66}  & \multicolumn{1}{l|}{21.8}                        & \multicolumn{1}{l|}{10.4}                         & \multicolumn{1}{l|}{14.1}                       & \multicolumn{1}{l|}{}    & \multicolumn{1}{l|}{\cellcolor[HTML]{EFEFEF}\textbf{20}}   & \multicolumn{1}{l|}{\cellcolor[HTML]{EFEFEF}\textbf{16.2}} & \multicolumn{1}{l|}{\cellcolor[HTML]{EFEFEF}\textbf{17.9}} & \multicolumn{1}{l|}{} & \multicolumn{1}{l|}{\textbf{85}}                 & \multicolumn{1}{l|}{\textbf{98.1}}                & \multicolumn{1}{l|}{\textbf{91.1}}              \\ \cline{1-5} \cline{7-9} \cline{11-13} 
\multicolumn{1}{|l|}{\cellcolor[HTML]{C0C0C0}darts}              & \multicolumn{1}{l|}{\cellcolor[HTML]{EFEFEF}75}  & \multicolumn{1}{l|}{100}                         & \multicolumn{1}{l|}{26.5}                         & \multicolumn{1}{l|}{41.7}                       & \multicolumn{1}{l|}{}    & \multicolumn{1}{l|}{\cellcolor[HTML]{EFEFEF}37}          & \multicolumn{1}{l|}{\cellcolor[HTML]{EFEFEF}\textbf{62.5}}          & \multicolumn{1}{l|}{\cellcolor[HTML]{EFEFEF}\textbf{46.5}}          & \multicolumn{1}{l|}{} & \multicolumn{1}{l|}{33.3}                        & \multicolumn{1}{l|}{\textbf{100}}                 & \multicolumn{1}{l|}{\textbf{50}}                \\ \cline{1-5} \cline{7-9} \cline{11-13} 
\multicolumn{1}{|l|}{\cellcolor[HTML]{C0C0C0}darts window}       & \multicolumn{1}{l|}{\cellcolor[HTML]{EFEFEF}50}  & \multicolumn{1}{l|}{25}                          & \multicolumn{1}{l|}{50}                           & \multicolumn{1}{l|}{33.3}                       & \multicolumn{1}{l|}{}    & \multicolumn{1}{l|}{\cellcolor[HTML]{EFEFEF}\textbf{33.3}} & \multicolumn{1}{l|}{\cellcolor[HTML]{EFEFEF}33.3}          & \multicolumn{1}{l|}{\cellcolor[HTML]{EFEFEF}33.3}          & \multicolumn{1}{l|}{} & \multicolumn{1}{l|}{\textbf{33.3}}               & \multicolumn{1}{l|}{33.3}                         & \multicolumn{1}{l|}{33.3}                       \\ \cline{1-5} \cline{7-9} \cline{11-13} 
\multicolumn{1}{|l|}{\cellcolor[HTML]{C0C0C0}softball}           & \multicolumn{1}{l|}{\cellcolor[HTML]{EFEFEF}96}  & \multicolumn{1}{l|}{66.7}                        & \multicolumn{1}{l|}{15.4}                         & \multicolumn{1}{l|}{25}                         & \multicolumn{1}{l|}{}    & \multicolumn{1}{l|}{\cellcolor[HTML]{EFEFEF}\textbf{83.3}} & \multicolumn{1}{l|}{\cellcolor[HTML]{EFEFEF}\textbf{83.3}} & \multicolumn{1}{l|}{\cellcolor[HTML]{EFEFEF}\textbf{83.3}} & \multicolumn{1}{l|}{} & \multicolumn{1}{l|}{54.5}                        & \multicolumn{1}{l|}{\textbf{66.7}}                & \multicolumn{1}{l|}{\textbf{60}}                \\ \cline{1-5} \cline{7-9} \cline{11-13} 
\multicolumn{1}{|l|}{\cellcolor[HTML]{C0C0C0}archery}            & \multicolumn{1}{l|}{\cellcolor[HTML]{EFEFEF}119} & \multicolumn{1}{l|}{0}                           & \multicolumn{1}{l|}{0}                            & \multicolumn{1}{l|}{0}                          & \multicolumn{1}{l|}{}    & \multicolumn{1}{l|}{\cellcolor[HTML]{EFEFEF}\textbf{25}}   & \multicolumn{1}{l|}{\cellcolor[HTML]{EFEFEF}\textbf{20}}   & \multicolumn{1}{l|}{\cellcolor[HTML]{EFEFEF}\textbf{22.2}} & \multicolumn{1}{l|}{} & \multicolumn{1}{l|}{\textbf{18.8}}               & \multicolumn{1}{l|}{\textbf{100}}                 & \multicolumn{1}{l|}{\textbf{31.6}}              \\ \cline{1-5} \cline{7-9} \cline{11-13} 
\multicolumn{1}{|l|}{\cellcolor[HTML]{C0C0C0}tennis serve side}  & \multicolumn{1}{l|}{\cellcolor[HTML]{EFEFEF}68}  & \multicolumn{1}{l|}{100}                         & \multicolumn{1}{l|}{58.8}                         & \multicolumn{1}{l|}{74.1}                       & \multicolumn{1}{l|}{}    & \multicolumn{1}{l|}{\cellcolor[HTML]{EFEFEF}66.7}          & \multicolumn{1}{l|}{\cellcolor[HTML]{EFEFEF}\textbf{76.9}} & \multicolumn{1}{l|}{\cellcolor[HTML]{EFEFEF}71.4}          & \multicolumn{1}{l|}{} & \multicolumn{1}{l|}{35.3}                        & \multicolumn{1}{l|}{\textbf{85.7}}                & \multicolumn{1}{l|}{50}                         \\ \cline{1-5} \cline{7-9} \cline{11-13} 
\multicolumn{1}{|l|}{\cellcolor[HTML]{C0C0C0}tennis serve back}  & \multicolumn{1}{l|}{\cellcolor[HTML]{EFEFEF}156} & \multicolumn{1}{l|}{28.6}                        & \multicolumn{1}{l|}{5.9}                          & \multicolumn{1}{l|}{9.8}                        & \multicolumn{1}{l|}{}    & \multicolumn{1}{l|}{\cellcolor[HTML]{EFEFEF}\textbf{35.3}} & \multicolumn{1}{l|}{\cellcolor[HTML]{EFEFEF}\textbf{69.2}} & \multicolumn{1}{l|}{\cellcolor[HTML]{EFEFEF}\textbf{46.8}} & \multicolumn{1}{l|}{} & \multicolumn{1}{l|}{26.4}                        & \multicolumn{1}{l|}{\textbf{70}}                  & \multicolumn{1}{l|}{\textbf{38.4}}              \\ \cline{1-5} \cline{7-9} \cline{11-13} 
\multicolumn{1}{|l|}{\cellcolor[HTML]{C0C0C0}tennis court}       & \multicolumn{1}{l|}{\cellcolor[HTML]{EFEFEF}128} & \multicolumn{1}{l|}{0}                           & \multicolumn{1}{l|}{0}                            & \multicolumn{1}{l|}{0}                          & \multicolumn{1}{l|}{}    & \multicolumn{1}{l|}{\cellcolor[HTML]{EFEFEF}\textbf{33.3}} & \multicolumn{1}{l|}{\cellcolor[HTML]{EFEFEF}\textbf{40.8}} & \multicolumn{1}{l|}{\cellcolor[HTML]{EFEFEF}\textbf{36.7}} & \multicolumn{1}{l|}{} & \multicolumn{1}{l|}{\textbf{25.5}}               & \multicolumn{1}{l|}{\textbf{58}}                  & \multicolumn{1}{l|}{\textbf{35.5}}              \\ \cline{1-5} \cline{7-9} \cline{11-13} 
\multicolumn{1}{|l|}{\cellcolor[HTML]{C0C0C0}hockey}             & \multicolumn{1}{l|}{\cellcolor[HTML]{EFEFEF}350} & \multicolumn{1}{l|}{100}                         & \multicolumn{1}{l|}{16.1}                         & \multicolumn{1}{l|}{27.7}                       & \multicolumn{1}{l|}{}    & \multicolumn{1}{l|}{\cellcolor[HTML]{EFEFEF}24.1}          & \multicolumn{1}{l|}{\cellcolor[HTML]{EFEFEF}\textbf{86.7}} & \multicolumn{1}{l|}{\cellcolor[HTML]{EFEFEF}\textbf{37.7}}          & \multicolumn{1}{l|}{} & \multicolumn{1}{l|}{20}                          & \multicolumn{1}{l|}{\textbf{91.7}}                & \multicolumn{1}{l|}{\textbf{32.8}}              \\ \cline{1-5} \cline{7-9} \cline{11-13} 
\multicolumn{1}{|l|}{\cellcolor[HTML]{C0C0C0}squash}             & \multicolumn{1}{l|}{\cellcolor[HTML]{EFEFEF}250} & \multicolumn{1}{l|}{0}                           & \multicolumn{1}{l|}{0}                            & \multicolumn{1}{l|}{0}                          & \multicolumn{1}{l|}{}    & \multicolumn{1}{l|}{\cellcolor[HTML]{EFEFEF}\textbf{26}}   & \multicolumn{1}{l|}{\cellcolor[HTML]{EFEFEF}\textbf{84.4}} & \multicolumn{1}{l|}{\cellcolor[HTML]{EFEFEF}\textbf{39.7}} & \multicolumn{1}{l|}{} & \multicolumn{1}{l|}{\textbf{21.6}}               & \multicolumn{1}{l|}{\textbf{75.9}}                & \multicolumn{1}{l|}{\textbf{33.6}}              \\ \cline{1-5} \cline{7-9} \cline{11-13} 
\multicolumn{1}{|l|}{\cellcolor[HTML]{C0C0C0}frisbee}            & \multicolumn{1}{l|}{\cellcolor[HTML]{EFEFEF}100} & \multicolumn{1}{l|}{100}                         & \multicolumn{1}{l|}{100}                          & \multicolumn{1}{l|}{100}                        & \multicolumn{1}{l|}{}    & \multicolumn{1}{l|}{\cellcolor[HTML]{EFEFEF}0}             & \multicolumn{1}{l|}{\cellcolor[HTML]{EFEFEF}0}             & \multicolumn{1}{l|}{\cellcolor[HTML]{EFEFEF}0}             & \multicolumn{1}{l|}{} & \multicolumn{1}{l|}{94.7}                        & \multicolumn{1}{l|}{94.7}                         & \multicolumn{1}{l|}{94.7}                       \\ \cline{1-5} \cline{7-9} \cline{11-13} 
\multicolumn{1}{|l|}{\cellcolor[HTML]{C0C0C0}blue ball}          & \multicolumn{1}{l|}{\cellcolor[HTML]{EFEFEF}53}  & \multicolumn{1}{l|}{100}                         & \multicolumn{1}{l|}{52.4}                         & \multicolumn{1}{l|}{68.8}                       & \multicolumn{1}{l|}{}    & \multicolumn{1}{l|}{\cellcolor[HTML]{EFEFEF}40}            & \multicolumn{1}{l|}{\cellcolor[HTML]{EFEFEF}26.7}          & \multicolumn{1}{l|}{\cellcolor[HTML]{EFEFEF}32}            & \multicolumn{1}{l|}{} & \multicolumn{1}{l|}{58.3}                        & \multicolumn{1}{l|}{\textbf{43.8}}                & \multicolumn{1}{l|}{50}                         \\ \cline{1-5} \cline{7-9} \cline{11-13} 
\multicolumn{1}{|l|}{\cellcolor[HTML]{C0C0C0}ping pong tampere}  & \multicolumn{1}{l|}{\cellcolor[HTML]{EFEFEF}120} & \multicolumn{1}{l|}{100}                         & \multicolumn{1}{l|}{88.7}                         & \multicolumn{1}{l|}{94}                         & \multicolumn{1}{l|}{}    & \multicolumn{1}{l|}{\cellcolor[HTML]{EFEFEF}58.6}          & \multicolumn{1}{l|}{\cellcolor[HTML]{EFEFEF}66.7}          & \multicolumn{1}{l|}{\cellcolor[HTML]{EFEFEF}62.4}          & \multicolumn{1}{l|}{} & \multicolumn{1}{l|}{0}                           & \multicolumn{1}{l|}{0}                            & \multicolumn{1}{l|}{0}                          \\ \cline{1-5} \cline{7-9} \cline{11-13} 
\multicolumn{1}{|l|}{\cellcolor[HTML]{C0C0C0}ping pong side}     & \multicolumn{1}{l|}{\cellcolor[HTML]{EFEFEF}445} & \multicolumn{1}{l|}{12.1}                        & \multicolumn{1}{l|}{7.3}                          & \multicolumn{1}{l|}{9.1}                        & \multicolumn{1}{l|}{}    & \multicolumn{1}{l|}{\cellcolor[HTML]{EFEFEF}\textbf{45.4}} & \multicolumn{1}{l|}{\cellcolor[HTML]{EFEFEF}\textbf{79.1}} & \multicolumn{1}{l|}{\cellcolor[HTML]{EFEFEF}\textbf{57.7}} & \multicolumn{1}{l|}{} & \multicolumn{1}{l|}{0}                           & \multicolumn{1}{l|}{0}                            & \multicolumn{1}{l|}{0}                          \\ \cline{1-5} \cline{7-9} \cline{11-13} 
\multicolumn{1}{|l|}{\cellcolor[HTML]{C0C0C0}ping pong top}      & \multicolumn{1}{l|}{\cellcolor[HTML]{EFEFEF}350} & \multicolumn{1}{l|}{92.6}                        & \multicolumn{1}{l|}{87.8}                         & \multicolumn{1}{l|}{90.1}                       & \multicolumn{1}{l|}{}    & \multicolumn{1}{l|}{\cellcolor[HTML]{EFEFEF}56}            & \multicolumn{1}{l|}{\cellcolor[HTML]{EFEFEF}\textbf{98.9}} & \multicolumn{1}{l|}{\cellcolor[HTML]{EFEFEF}71.5}          & \multicolumn{1}{l|}{} & \multicolumn{1}{l|}{0}                           & \multicolumn{1}{l|}{0}                            & \multicolumn{1}{l|}{0}                          \\ \cline{1-5} \cline{7-9} \cline{11-13} 
\cellcolor[HTML]{FFFFFF}Average per frame                        & \cellcolor[HTML]{FFFFFF}2476                     & \cellcolor[HTML]{FFFFFF}53.7                     & \cellcolor[HTML]{FFFFFF}31                        & \cellcolor[HTML]{FFFFFF}35.5                    & \cellcolor[HTML]{FFFFFF} & \cellcolor[HTML]{FFFFFF}38.3                      & \cellcolor[HTML]{FFFFFF}\textbf{68.5}                        & \cellcolor[HTML]{FFFFFF}\textbf{47.2}                      &                       & 21.7                                             & \textbf{49.7}                                     & 27.8                                           
\end{tabular}
}
\caption{Performance of the original CVPR2017 method \cite{rozumnyi2017world} in comparison to proposed method (method I - trained for smaller foregrounds; method II - trained for bigger foregrounds). The results suggests better overall performance of the trace segmentation in overall F1 performance score for method I.}
\label{tab:results}
\end{table*}

The performance of the method reflects the purpose of our algorithm. It performs well on sequences with small ball-shaped object moving relatively fast (ping-pong, softball, tennis and squash). Poor performance was recorded on sequences with foregrounds different from balls (like darts or archery) and on sequences with low background-foreground contrast (darts window and blue ball). The method under-performs on data with grater foreground / velocity ratio (frisbee and volleyball). The foreground on these sequences are of larger size and is not moving faster than its diameter, as per FMO definition in Section \ref{sec:intro}.

Our approach is advantageous in fact that the network can be easily fine-tuned with image synthesizer setup for another sequence type, such as particular background, particular foreground (i.e. yellow ball) or foregrounds of different size range, etc. For comparison we have re-trained the network to detect foreground of bigger size and slower motions. The results are in the most right part of the results table \ref{tab:results}. The segmentation network stopped to be sensitive to smaller foregrounds, such as ping pong, squash or tennis, and starts to perform in cases with larger foregrounds, like frisbee or volleyball.

The next main difference is that our algorithm tackle the problem of detecting very small objects (from roughly 2 pixels in diameter) or objects crossing the background of similar color.


\begin{table}[]
\captionsetup{font=scriptsize}
\centering{
\begin{tabular}{cc}
\multicolumn{1}{l}{\textbf{video resolution}}      & \multicolumn{1}{l}{\textbf{average fps}} \\ \hline
\multicolumn{1}{|c|}{\textbf{864 x 1536}} & \multicolumn{1}{c|}{2}                   \\ \hline
\multicolumn{1}{|c|}{\textbf{576 x 1024}} & \multicolumn{1}{c|}{4.7}                 \\ \hline
\multicolumn{1}{|c|}{\textbf{430 x 768}}  & \multicolumn{1}{c|}{8.6}                 \\ \hline
\multicolumn{1}{|c|}{\textbf{324 x 576}}  & \multicolumn{1}{c|}{11.8}                \\ \hline
\multicolumn{1}{|c|}{\textbf{216 x 384}}  & \multicolumn{1}{c|}{23.1}                \\ \hline
\end{tabular}
}
\caption{Some examples of video inference times achieved using NVidia Tesla X GPU.}
\label{tab:performance}
\end{table}

\subsection{Computational time}

Another benefit that comes from using the neural network is relatively short inference time. The state-of-the-art approaches\cite{rozumnyi2019non,kotera2018motion,kotera2019intra} are based on foreground de-blurring and therefore are inherently slow. In \cite{kotera2019intra} authors state mean time is 4 second per frame. 
Our methods is capable of running in near-realtime regime using widely available graphics card. For more details please refer to Table \ref{tab:performance}, where we depicted mean frame evaluation times for NVidia Tesla X GPU using several samples of image resolutions.

Next goal will be to speed up the inference times enough for the method to be able to perform in real-time environment. We plan to achieve this by optimizing the network in size by pruning, using compacted backbone, reducing precision or network quantization.

\subsection{YouTube sport videos}

\begin{figure*}[t]
  \centering
  \includegraphics[width=0.75\textwidth]{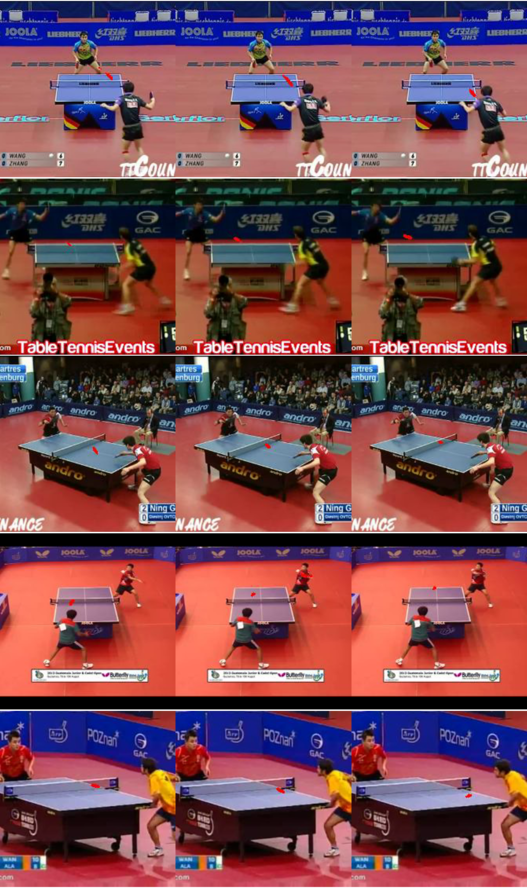}
\caption{YouTube real world ping pong sequences evaluation. }
\label{fig:pingpong}
\end{figure*}

As mentioned before, our primary goal was tracking of balls in sporting videos with predominately high relative speed, such as ping pong, baseball, tennis or badminton. For this purpose we have downloaded more than 900 000 YouTube sport videos to create a base of our synthetic data generator backgrounds. Over 1800 of this sequences contain ping-pong matches, which we used for testing of our framework. Although we measured our performance on the FMO dataset, we also aim for good performance on real wold sequences. Examples of ping-pong sequence evaluation can be seen in Figure \ref{fig:pingpong}.

\section{Conclusion}
We proposed a method performing in difficult task of real time real world fast moving object detection and tracking. 
We achieved to overcome limitations of the previous works in this field, namely the long computation time and difficulty to detect small and very fast objects or objects crossing the background of similar color. We have introduced a synthetic physically plausible fast moving object sequence generator, which we utilize for network training. We showed the simplicity of adapting the generator to another type of foreground followed by network fine-tuning that allows us to detect foregrounds of different size and color.

In the future work, we would like to focus on optimizing the processing pipeline with respect to speed in order to achieve true real-time performance in high resolution videos and automatically track all kinds of sports balls in video streams. This can be further utilized in various applications such as instantaneous ball speed detection, ball misses or ball out of bounds detection.




\listoftables



\bibliographystyle{IEEEtran}
%
\bibliography{references}



\end{document}